\documentclass[a4paper,fleqn]{cas-dc}
\usepackage{latexsym}
\usepackage{bm}
\usepackage{amssymb} 
\usepackage{array}
\usepackage{amsmath}
\usepackage{siunitx}
\usepackage[authoryear,longnamesfirst]{natbib}

\def\tsc#1{\csdef{#1}{\textsc{\lowercase{#1}}\xspace}}
\tsc{WGM}
\tsc{QE}
\tsc{EP}
\tsc{PMS}
\tsc{BEC}
\tsc{DE}

\begin{document}
\let\WriteBookmarks\relax
\def\floatpagepagefraction{1}
\def\textpagefraction{.001}

\shorttitle{Prostate Segmentation on micro-US Images}

\shortauthors{Jiang et~al.}

\title [mode = title]{MicroSegNet: A Deep Learning Approach for Prostate Segmentation on Micro-Ultrasound Images}             

\author[1]{Hongxu Jiang}
\credit{Methodology, Software, Formal analysis, Investigation, Data curation, Writing - original draft, Visualization}
\fnmark[1]
\author[2]{Muhammad Imran}
\credit{Methodology, Data curation, Writing - original draft, Visualization}
\fnmark[1]
\author[3]{Preethika Muralidharan}
\credit{Data curation, Writing - review \& editing}
\author[4]{Anjali Patel}
\credit{Data curation, Writing - review \& editing}
\author[5]{Jake Pensa}
\credit{Data curation, Writing - review \& editing}
\author[6]{Muxuan Liang}
\credit{Formal analysis, Writing - review \& editing}
\author[7]{Tarik Benidir}
\credit{Writing - review \& editing}
\author[8]{Joseph R. Grajo}
\credit{Writing - review \& editing}
\author[7]{Jason P. Joseph}
\credit{ Writing - review \& editing}
\author[7]{Russell Terry}
\credit{ Writing - review \& editing}
\author[7]{John Michael DiBianco}
\credit{Data curation, Writing - review \& editing}
\author[7]{Li-Ming Su}
\credit{Supervision, Writing - review \& editing}
\author[9]{Yuyin Zhou}
\fnmark[2]
\credit{Methodology, Supervision, Writing - original draft}
\author[10]{Wayne G. Brisbane}
\fnmark[2]
\credit{Conceptualization, Resources, Supervision, Writing - original draft, Project administration}
\author[2]{Wei Shao}[
orcid = 0000-0003-4931-4839]
\fnmark[2]
\cormark[1]
\ead{weishao@ufl.edu}
\credit{Conceptualization, Methodology, Resources, Data curation, Writing - original draft, Supervision, Project administration, Funding acquisition}
\affiliation[1]{organization={Department of Electrical and Computer Engineering, University of Florida},
    city={Gainesville},
    state={Florida},
    postcode={32608}, 
    country={United States}
    }

    \affiliation[2]{organization={Department of Medicine, University of Florida},
    city={Gainesville},
    state={Florida},
    postcode={32608}, 
    country={United States}
    }

    \affiliation[3]{organization={Department of Health Outcomes and Biomedical Informatics, University of Florida},
    city={Gainesville},
    state={Florida},
    postcode={32608}, 
    country={United States}
    }

\affiliation[4]{organization={College of Medicine , University of Florida},
    city={Gainesville},
    state={Florida},
    postcode={32608}, 
    country={United States}
    }

\affiliation[5]{organization={Department of  Bioengineering, University of California},
    city={Los Angeles},
    state={California},
    postcode={90095}, 
    country={United States}
    }

\affiliation[6]{organization={Department of Biostatistics, University of Florida},
   city={Gainesville},
    state={Florida},
    postcode={32608}, 
    country={United States}
    }

\affiliation[7]{organization={Department of Urology, University of Florida},
    city={Gainesville},
    state={Florida},
    postcode={32608}, 
    country={United States}
    }

\affiliation[8]{organization={Department of Radiology, University of Florida},
    city={Gainesville},
    state={Florida},
    postcode={32608}, 
    country={United States}
    }

\affiliation[9]{organization={Department of Computer Science and Engineering, University of California},
    city={Santa Cruz},
    state={California},
    postcode={95064}, 
    country={United States}
    }

\affiliation[10]{organization={Department of  Urology, University of California},
    city={Los Angeles},
    state={California},
    postcode={90095}, 
    country={United States}
    }

\cortext[cor1]{Corresponding author}
\fntext[fn1]{Equal contribution as first author}
\fntext[fn2]{Equal contribution as senior author}

\begin{abstract}
Micro-ultrasound (micro-US) is a novel 29-MHz ultrasound technique that provides 3-4 times higher resolution than traditional ultrasound, potentially enabling low-cost, accurate diagnosis of prostate cancer.
Accurate prostate segmentation is crucial for prostate volume measurement, cancer diagnosis, prostate biopsy, and treatment planning. 
However, prostate segmentation on micro-US is challenging due to artifacts and indistinct borders between the prostate, bladder, and urethra in the midline. 
This paper presents MicroSegNet, a multi-scale annotation-guided transformer UNet model designed specifically to tackle these challenges.
During the training process, MicroSegNet focuses more on regions that are hard to segment (hard regions), characterized by discrepancies between expert and non-expert annotations.
We achieve this by proposing an annotation-guided binary cross entropy (AG-BCE) loss that assigns a larger weight to prediction errors in hard regions and a lower weight to prediction errors in easy regions. 
The AG-BCE loss was seamlessly integrated into the training process through the utilization of multi-scale deep supervision, enabling MicroSegNet to capture global contextual dependencies and local information at various scales.
We trained our model using micro-US images from 55 patients, followed by evaluation on 20 patients.
Our MicroSegNet model achieved a Dice coefficient of 0.939 and a Hausdorff distance of 2.02 mm, outperforming several state-of-the-art segmentation methods, as well as three human annotators with different experience levels.
Our code is publicly available at \url{https://github.com/mirthAI/MicroSegNet} and our dataset is publicly available at \url{ https://zenodo.org/records/10475293}.

\end{abstract}


\maketitle

\section{Introduction}
\label{sect:introduction}
Prostate cancer is the most frequently diagnosed cancer in the United States, with an estimated 268,490 new cases in 2022, and it is the second leading cause of cancer-related deaths among men in the United States, with an estimated 34,500 deaths in the same year \citep{siegel2022cancer}. 
Early diagnosis of clinically significant prostate cancer can improve the 5-year survival rate from 31\% to nearly 100\% \citep{wolf2010american}. 
In the United States, approximately 1 million transrectal ultrasound (TRUS) guided prostate biopsy procedures are performed annually \citep{loeb2011complications}.  Due to the poor accuracy of the TRUS imaging systems, many guideline organizations suggest performing both targeted and blind systematic biopsies \citep{mottet2021eau}.  As a result only 12\% of biopsy cores uncover clinically significant cancer \citep{brisbane2022targeted}, causing bleeding, infection, and urinary difficulties related to unnecessary biopsies.

Micro-ultrasound (micro-US) is a novel 29-MHz ultrasound technology that offers 3-4 times higher resolution than traditional ultrasound. 
While the efficacy of micro-ultrasound in prostate cancer diagnosis is currently a matter of debate, emerging studies suggest it may have the potential to achieve diagnostic accuracy comparable to MRI at a lower cost \citep{klotz2021comparison,sountoulides2021micro,dias2022multiparametric}.
One notable advantage of micro-US is its real-time visualization capability, which eliminates the need for image fusion, a process that can contribute up to 50\% of diagnostic errors in MRI-ultrasound fusion-guided biopsy \citep{williams2022does}. 
However, complexities in image interpretation limit the widespread adoption of micro-US \citep{pavlovich2021multi}. 
Machine learning algorithms assist clinicians with image interpretation and identification of cancer; however, machine learning for micro-US is immature.  The first step in diagnostic machine learning is the ability to automatically identify the prostate capsule.  Once identified, the capsule will allow for three important advancements:  First, a defined prostate volume will enable registration of micro-US to any other diagnostic volume including histopathology, MRI, prostate-specific membrane antigen (PSMA), and biopsy coordinates. Second, cancer staging is based on tumor location relative to the prostate capsule.  Finally, the ability of machine learning to identify the prostate capsule will serve as a minimum quality benchmark for cancer evaluation.

However, prostate segmentation on micro-US poses cognitive challenges for three main reasons.
First, urologists and radiologists are accustomed to examining prostate images in the axial plane (MRI and CT), whereas micro-US provides oblique planes. Second, in the midline, there exist indistinct borders between the prostate, bladder, and surrounding vasculature. Third, image artifacts can arise on micro-US due to prostate diseases like prostate calcification.
Consequently, micro-US tracks the locations of all biopsy samples relative to the urethra (midline) rather than relative to the capsule. 
This results in a time-consuming process of manual prostate segmentation, which renders high-volume prostate segmentation impractical for clinical practice.
Despite the improved accuracy offered by micro-US, many urologists still prefer MRI for treatment planning since it can generate a 3-dimensional (3D) cancer map. 
Therefore, precise and scalable capsule segmentation is crucial to accurately determine the location of prostate cancer in relation to the rest of the prostate. 

In this paper, we introduce MicroSegNet, the first deep learning based approach for prostate segmentation on micro-US images. 
MicroSegNet is built upon the powerful TransUNet architecture \citep{chen2021transunet}, which incorporates a self-attention mechanism to enhance feature representation, resulting in more accurate detection of the boundary and refined segmentation results. 
To capture finer details, we  integrate multi-scale deep supervision \citep{xie2015holistically}into TransUNet, enabling the model to effectively capture both global contextual dependencies and local information across different scales. 
To address challenges in segmenting hard-to-segment regions (hard regions), we propose an innovative annotation-guided cross entropy (AG-BCE) loss function.
Unlike existing prostate segmentation methods that solely utilize expert annotations in the loss functions, our study is the first to exploit the benefits of non-expert annotations in improving prostate segmentation.
Our AG-BCE loss function assigns higher penalties to prediction errors in hard regions compared to easy regions, encouraging the model to focus on accurately segmenting these challenging areas. 
Hard and easy regions are identified based on the agreement or disagreement observed between expert and non-expert annotations, with hard regions indicating higher uncertainty and complexity. 
To encourage further research that utilizes non-expert annotations, we will make these annotations publicly available as part of our dataset.
 
By incorporating multi-scale deep supervision and the AG-BCE loss into the TransUNet architecture, MicroSegNet achieves significant improvements in prostate segmentation performance, particularly in hard-to-segment regions. The model effectively allocates attention and resources to the hard regions, leading to better alignment with expert annotations and enhanced accuracy in segmenting the prostate on micro-US images.
We summarize the major contributions of this paper as the following:
\begin{itemize}
\item We introduced MicroSegNet, the first deep learning model for automated prostate segmentation on micro-ultrasound images. 
\item We proposed a novel annotation-guided cross entropy loss which assigns higher penalties to prediction errors in difficult-to-segment regions, allowing the model to focus on these challenging areas during the training.
\item We curated the first micro-US dataset comprising both expert and non-expert prostate annotations of 75 patients, serving as a valuable resource for training and validating prostate segmentation models on micro-US images.
\end{itemize}

\section{Related Works}
\label{sect:relatedwork}
\subsection{Deep learning for prostate segmentation}

Accurate segmentation of the prostate capsule is crucial for various clinical applications, including prostate cancer diagnosis and treatment planning. In recent years, significant advances have been made in developing robust and accurate prostate segmentation techniques for various imaging modalities, including MRI \citep{jia2017prostate, soerensen2021prognet,ghavami2019automatic, chen2021medical}, and TRUS \citep{lei2019ultrasound, lei2019ultrasound_2, zhu2019boundary, 8698868, https://doi.org/10.1002/mp.14134, peng2023automatic, peng2022h}. Our MicroSegNet model focuses on prostate segmentation on micro-US images, particularly aligning with TRUS-based prostate segmentation techniques. For example, Wang et al. \citep{8698868} introduced an attention-based convolutional neural network (CNN) model to enhance prostate details by suppressing non-prostate noise. Karimi et al. \citep{KARIMI2019186} developed a model trained separately on simple and complex ultrasound scans, using their differences to achieve improved segmentation. Tao Peng et al. \citep{ PENG2022106752} proposed H-ProSeg, a hybrid prostate segmentation algorithm that combines an improved principal curve method with machine learning, introducing techniques like RCKPC and MCSDE to enhance accuracy and generate smooth prostate contours. Orlando et al. \citep{https://doi.org/10.1002/mp.14134} enhanced the UNet architecture by incorporating additional dropout and transposed convolution layers. A hybrid method, H-SegMed [13], combines an enhanced closed principal curve-based approach with machine learning to produce accurate, smooth prostate contours. Vesal et al. \citep{vesal2022domain} improved the UNet architecture by integrating residual connections and dilated convolutions for detailed contextual information capture. In a separate study \citep{peng2023automatic}, a two-stage prostate segmentation method was introduced, using an attention-based U-Net architecture for initial contour estimation and a modified principal curve-based approach with an evolutionary neural network for refinement.

Prior conventional ultrasound prostate segmentation methods were mainly designed for 3D ultrasound images, addressing challenges like blurred prostate boundaries and low image contrast due to low signal-to-noise ratios \citep{lei2019ultrasound,wang2023prostate}, large intensity variations with the prostate \citep{yang2017fine}, and shadow artifacts \citep{xu2021shadow,vesal2022domain}. Micro-ultrasound segmentation faces issues such as unclear prostate borders and artifacts from prostate calcification. The key differences between conventional and micro-ultrasound prostate segmentation are: (1) conventional methods focus on 3D volume post-probe sweep, lacking real-time capabilities, whereas MicroSegNet works in real-time on micro-ultrasound images in pseudo-sagittal view. (2) micro-ultrasound's resolution is 3-4 times higher than conventional ultrasound, making the primary challenge the imaging artifacts from calcification, despite its higher image contrast.

\subsection{Loss functions in image segmentation}
Image segmentation methods often utilize cross entropy as the loss function, which tends to favor regional information over boundary details. To address this issue, researchers have developed methods to enhance the cross entropy loss function.
In the U-Net paper, Ronneberger et al. \citep{ronneberger2015u} proposed a weighted cross-entropy loss that employs a weight map based on the distance to cell boundaries for learning the small separation borders between cells. 
More recently, several boundary-aware segmentation losses based on the distance transform map have been proposed. 
Zhu et al. \citep{zhu2019boundary} introduced a boundary-weighted segmentation loss (BWSL) function to make the segmentation network more sensitive to boundaries during segmentation. Kervadec et al. \citep{kervadec2019boundary} proposed a boundary loss formulated as a distance metric focusing on contours rather than regions. This approach addresses unbalanced segmentation challenges by emphasizing integrals over boundaries instead of regions. Karimi et al. \citep{karimi2019reducing} proposed three loss functions based on the distance transform of the segmentation boundary, aimed at minimizing the Hausdorff distance in medical image segmentation. Ma et al. \citep{ma2020distance} conducted a comprehensive comparison of several boundary-aware segmentation losses across different datasets. 
In contrast to previous boundary-aware segmentation losses that uniformly address organ boundaries, our AG-BCE loss uniquely employs a weighting scheme based on the consensus between expert and non-expert annotations, enabling our MicroSegNet model to target challenging prostate boundaries more effectively.

\subsection{Transformers in medical image segmentation} 
The transformer architecture, originally introduced by Vaswani et al. \citep{vaswani2017attention} for natural language processing tasks has achieved remarkable success in machine translation, text classification, and language modeling. Its self-attention mechanism allows for capturing contextual information between positions in the input sequence. Recently, transformer-based architectures have shown promising performance in medical image segmentation \citep{chen2021transunet,hatamizadeh2022unetr,xie2021cotr}.
In the context of prostate segmentation, various approaches have been proposed. 
Wang et al. \citep{wang2022multiscale} introduced a multiscale fusion scheme using transformers, while Zhang et al. \citep{zhang2021transfuse} combined transformers and CNNs in a parallel-in-branch architecture called TransFuse. 
CSA \citep{ding2023multi} leveraged multi-scale and channel-wise self-attention, and CAT-Net~\citep{hung2022cat} incorporated a cross-slice attention mechanism within a transformer module. Our study utilizes TransUNet~\citep{chen2021transunet}, the first transformer-based architecture designed for medical image segmentation.

\section{Methods}
\label{sect:methodology}
\subsection{Image acquisition}
\label{sect:imageacquisition}
\begin{figure}[!hbt]
    \centering
    \includegraphics[width=0.48\textwidth]{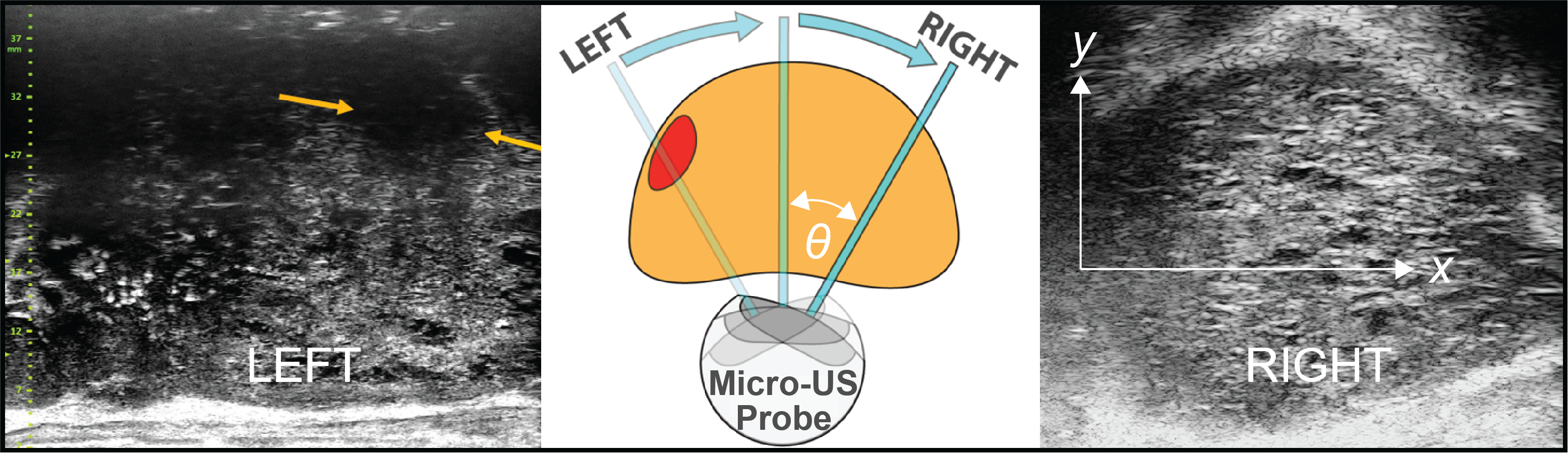}
    \caption{An example of micro-US scanning.  The micro-US probe is rotated under the prostate (yellow) to generate serial 2-dimensional images separated by a rotational angle $\theta$. This prostate also has a visible cancer lesion (red region and yellow arrows)}
    \label{fig:Example of micro-US scanning}
\end{figure}

This study, approved by the institutional review board at the University of Florida, included 75 men who underwent micro-US-guided prostate biopsy at the University of Florida between September 2021 and December 2022.
Patients had a clinical indication for prostate biopsy including 1) PSA elevation, 2) biomarker abnormality, and/or 3) abnormal MRI. 
Prior to biopsy, patients underwent a transrectal rotational scan of the prostate from left to right capturing approximately 200 to 300 micro-US images.
During the prostate scan, the probe was kept stationary in the cranial-to-caudal orientation by hand. However, it is important to note that the probe is constrained from lateral motion (left/right) by the rectal sphincter. The cranial-to-caudal orientation is limited by the natural curve of the sigmoid colon, preventing advancement past the seminal vesicles. Furthermore, the probe's design features a gentle edge that naturally sits just distal to the rectal sphincter, preventing significant caudal movement. An apparatus (Civco brachytherapy arm) was initially used but was abandoned as it poorly contacted the prostate throughout the entire range of image acquisition.
The captured images contained pixels in the pseudo-sagittal plane separated by the rotation angle $\theta$ (see Figure \ref{fig:Example of micro-US scanning}).
Following scan acquisition, the data was imported into a customized MATLAB script to convert the b-mode image set into a DICOM series with embedded pixel spacing information. 
The recorded rotation angle of each image was stored as the through plane slice location to allow for angle identification during review of the sweep.

\subsection{Prostate annotation}
\label{sect:prostateannotation}
Manual prostate capsule segmentations were performed on micro-US images of 75 patients by two non-expert annotators and one expert annotator.
The first non-expert annotator (HJ) is a beginner annotator who had six months of experience in micro-US. 
The second non-expert annotator (WS) is an intermediate annotator who had four years of experience in prostate imaging and six months of experience in micro-US. 
The non-expert annotators received training from an expert urologist (WB) who had seven years of experience in prostate imaging and four years in interpreting micro-ultrasound images.
The expert provided ground-truth annotations for 10 independent micro-US scans, facilitating a detailed exploration of prostate anatomy and techniques for accurately delineating its borders in the images. The annotators practiced on these 10 micro-US scans, refining their skills by systematically comparing their annotations with the expert's ground-truth annotations.

Using the open-source 3D Slicer software \citep{fedorov20123d}, the beginner annotator (HJ) annotated the prostate capsule on every four micro-US slices, resulting in 50 to 60 segmented 2D slices for each of the 75 micro-US sweeps. 
The intermediate annotator (WS) subsequently  updated the annotations made by the beginner annotator. 
Finally, the expert urologist (WB) reviewed the annotations from the intermediate annotator and provided feedback and suggestions for improvement.
A review of the ultrasound by the expert took approximately 45 minutes. Each ultrasound slice was individually reviewed. If the non-expert annotations were off, they were edited by the expert. If satisfactory, they were left unchanged.
The non-expert annotators then worked together to refine their annotations accordingly until they reached a consensus with the expert annotator. 
The final annotations were considered as the expert-level prostate segmentations for this study. 
The initial annotations by the beginner annotator were considered as the non-expert prostate segmentations.
The annotation process took approximately four hours per micro-US scan, including two hours for the initial annotation and two hours for the refinement based on the expert urologist's feedback.

To assess the accuracy of our deep learning model compared to human annotators, three additional annotators manually segmented the prostate capsule on 20 testing cases using 3D Slicer.
The first annotator (AP) was a first-year medical student who underwent 28 h of training.
The second annotator (PM) was a first-year master's student who received 80 h of training.
The third annotator (JD) was a urologist with 1 year of experience in reading micro-US images.

\subsection{Proposed MicroSegNet model}
\label{sect:prostatesegmodel}
\begin{figure*}[!hbt]
\centering
\includegraphics[width=1\textwidth]{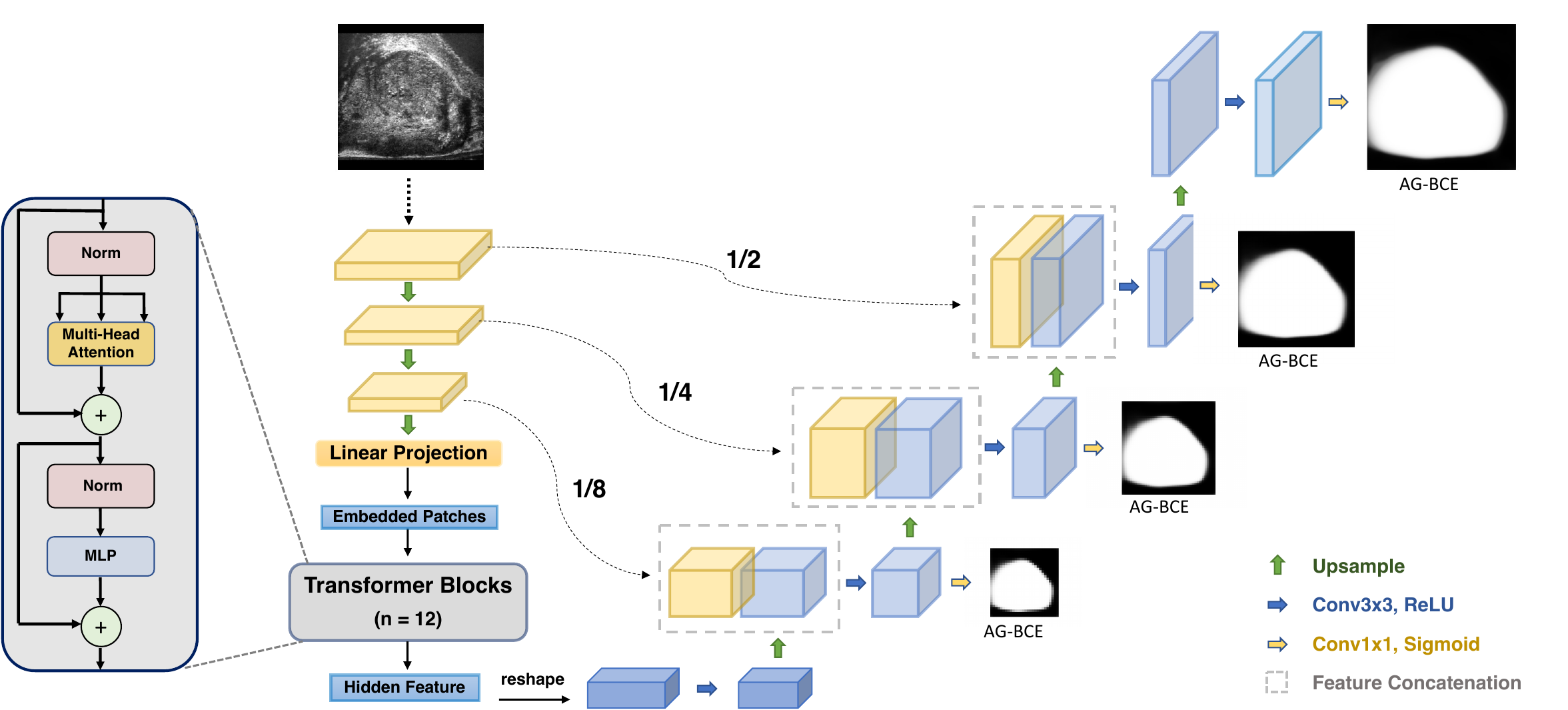}
\caption{Overview of the proposal MicroSegNet model. Based on the TransUNet architecture, the model enhances global context capture by incorporating the multi-scale deep supervision module in intermediate layers. The AG-BCE loss in the final output layer focuses on improving segmentation accuracy in difficult areas. }
\label{fig:framework}
\end{figure*}

Figure~\ref{fig:framework} provides an overview of the proposed MicroSegNet model, which is based on the TransUNet architecture, a transformer-based model for medical image segmentation. To enhance the performance of TransUNet, we implemented two key improvements.
First, we incorporated a multi-scale deep supervision module that facilitates the capture of global contextual dependencies while retaining local information.
Second, we introduced the annotated-guided AG-BCE loss, which we integrated into the final output layer and all intermediate layers. This novel loss function allowed MicroSegNet to prioritize the accurate segmentation of challenging regions during the training process. The combination of the multi-scale deep supervision module and the AG-BCE loss significantly boosted MicroSegNet's segmentation capabilities.

\subsubsection{TransUNet architecture}

The Vision Transformer (ViT), introduced by Dosovitskiy et al. \citep{dosovitskiy2020image}, is a transformer-based architecture originally designed for image classification tasks. 
It utilizes tokenized image patches and employs the self-attention mechanism \citep{vaswani2017attention} to effectively capture global context. 
TransUNet is the first transformer-based medical image segmentation model that combines the strengths of ViT and the UNet architectures. It harnesses the power of self-attention mechanisms and a hybrid CNN-transformer design to capture both global context and high-resolution spatial information effectively. This enables TransUNet to achieve precise localization by integrating self-attentive features with high-resolution CNN features. By combining the global context modeling capabilities of transformers with the accurate segmentation abilities of the UNet framework, TransUNet delivers superior segmentation results than previous CNN-based models.
Technically, TransUNet consists of the following components:
\begin{itemize}
    \item \textbf{Convolutional stem:} A convolutional stem layer extracts low-level features from the input image and reduces its spatial resolution.
    \item \textbf{Input sequentialization:} 
    Following the approach by \citep{dosovitskiy2020image}, we employ tokenization to convert the input features $\bm{\mathrm{x}}$ 
    obtained from the convolutional stem into a sequential representation of flattened 2D patches denoted as
    ${\bm{\mathrm{x}}^i_p \in \mathbb{R}^{P^2 \cdot C}|i=1,..,N}$. 
    Each patch has dimensions $P \times P$, and the total number of image patches is determined by $N=\frac{HW}{P^2}$, where $H$ and $W$ represent the height and width of the input image, respectively. This sequentialization process transforms the input into an input sequence of fixed length.
    \item \textbf{Patch embedding:} A patch embedding layer splits the feature map into fixed-size patches $\bm{\mathrm{x}}_p$ and projects them into a $D$-dimensional embedding space using a linear transformation. 
    The patch embeddings are then added with positional embeddings that encode the spatial information of the patches, which can be written as:
\begin{align}
    \bm{\mathrm{z}}_0 &= [\bm{\mathrm{x}}^1_p \bm{\mathrm{E}}; \, \bm{\mathrm{x}}^2_p \bm{\mathrm{E}}; \cdots; \, \bm{\mathrm{x}}^{N}_p \bm{\mathrm{E}} ] + \bm{\mathrm{E}}_{pos}, \label{eq:embedding} 
\end{align}
\noindent where $\bm{\mathrm{E}} \in \mathbb{R}^{(P^2 \cdot C) \times D}$ is the patch embedding projection, and $\bm{\mathrm{E}}_{pos}  \in \mathbb{R}^{N \times D}$ denotes the position embedding. 

        \item \textbf{Transformer encoder:} A transformer encoder that consists of $L$ layers of multi-head self-attention (MHSA) and feed-forward networks (FFN) (Eq.~\eqref{eq:MHSA_apply}\eqref{eq:FFN_apply}), with residual connections and layer-wise normalization. The transformer encoder receives the patch embeddings as input and updates them iteratively by attending to different patches based on their relevance. 
     The output of the $\ell$-th layer can be written as follows:
    \begin{align}
        \bm{\mathrm{z}}^\prime_\ell &= MHSA(LN(\bm{\mathrm{z}}_{\ell-1})) + \bm{\mathrm{z}}_{\ell-1}, &&  \label{eq:MHSA_apply} \\
        \bm{\mathrm{z}}_\ell &= FFN(LN(\bm{\mathrm{z}}^\prime_{\ell})) + \bm{\mathrm{z}}^\prime_{\ell},   \label{eq:FFN_apply} 
    \end{align}
    where $LN(\cdot)$ denotes the layer-wise normalization operator and $\bm{\mathrm{z}}_L$ is the encoded image representation. 
    \item \textbf{UNet Decoder:} A UNet decoder consists of several up-sampling blocks, each with a convolutional layer, an up-sampling layer, and a skip connection from the corresponding convolutional stem layer. The UNet decoder receives the patch embeddings as input and up-samples them to match the original image resolution. The skip connections help to fuse the high-resolution features from the convolutional stem with the high-level features from the transformer encoder. The output of the UNet decoder is a segmentation map that assigns a class label to each pixel in the input image.
\end{itemize}


\subsubsection{Multi-scale deep supervision}
The TransUNet model, as a baseline, aims to minimize the discrepancy between the predicted segmentation map and the ground truth segmentation. However, this approach may lack robustness when faced with variations in the shape, size, and appearance of input images. To address these challenges, we employed multi-scale deep supervision, a technique that incorporates information from multiple scales during the segmentation process \citep{xie2015holistically}.

Each pathway in the network captures and analyzes features at a specific scale, enabling the model to consider both local and global  information. In our study, we introduced 1 $\times$ 1 convolutional layers with the Sigmoid activation function to intermediate image features at three different scales ($\frac{1}{2}$, $\frac{1}{4}$, and $\frac{1}{8}$). This resulted in prostate segmentation images at varying resolutions (refer to Figure \ref{fig:framework}), where lower-scale outputs encompassed global information, while higher-scale outputs emphasized local details.
To ensure accurate predictions across different scales, we downsampled the ground truth segmentation and incorporated the discrepancy between each prediction and its corresponding downsampled ground truth segmentation image into the loss function. By including supervision for predictions at multiple scales, our model seamlessly integrated diverse feature levels, allowing for the capture of both high-level and local contextual information.

\subsubsection{Annotation-guided binary cross entropy loss}
\label{sect:AG-BCEloss}

\noindent\textbf{Definition of hard and easy regions.}
We have observed that accurately segmenting certain regions of the prostate (e.g., the borders between the prostate and the bladder in the midline) presents challenges even for experienced urologists. In order to overcome these difficulties and improve prostate segmentation, our focus lies on enhancing the accuracy of segmentation in these challenging regions, which we refer to as "hard regions."

\begin{figure}[!h]
\centering
\includegraphics[width=0.33\textwidth]{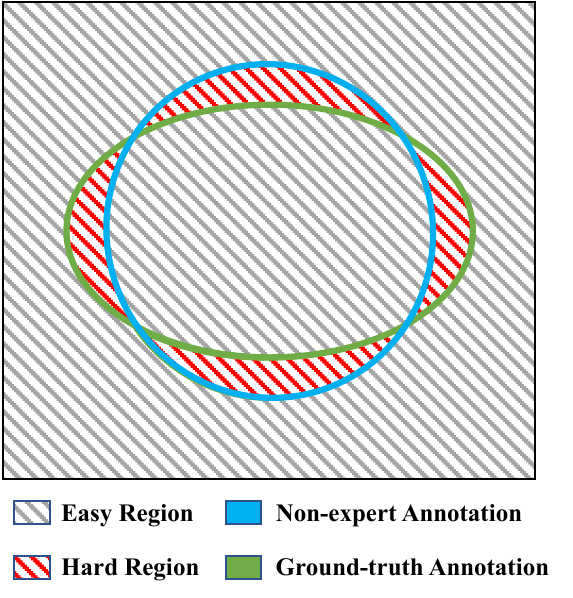}
\caption{Definition of hard and easy regions based on expert and non-expert annotators.}
\label{fig:hard_easy_regions}
\end{figure}

Manually delineating the hard and easy regions on each micro-US image is a labor-intensive task and can be subjective. To address this, we leveraged the expert and non-expert annotations obtained during the prostate annotation process, eliminating the need for additional manual effort. By incorporating non-expert annotations, which are generally more cost-effective to obtain, we defined hard regions as areas where discrepancies exist between the expert and non-expert annotations (as depicted in Figure~\ref{fig:hard_easy_regions}). Conversely, easy regions were identified as areas where there was agreement or overlap between the expert and non-expert annotations.

To address  segmentation challenges in the hard regions, we propose a novel AG-BCE loss. This innovative loss function allows the model to prioritize the hard regions over the easier-to-segment regions during the training process. By utilizing this annotation-guided approach, our model effectively focuses its attention on the challenging regions of the prostate, leading to improved segmentation performance in those areas.

\noindent\textbf{AG-BCE loss function.}
One of the most commonly used loss functions for binary image segmentation is the BCE loss, which measures the dissimilarity between the predicted probability map $P$ and the ground truth segmentation $Y$. 
The BCE loss is defined as follows:
\begin{equation}
    BCE(P, Y) = -\frac{1}{K}\sum_{i = 1}^{K} \left( y_i\cdot \log p_i + (1-y_i)\cdot \log (1-p_i) \right)
\end{equation}
where $K$ is the number of pixels in the image, $y_i$ is the ground truth label for pixel $i$, and $p_i$ is the predicted probability for pixel $i$. 
The BCE loss assigns equal weight to all pixels in the image, implying that all pixels are equally difficult to segment. 
However, this assumption does not hold for prostate segmentation on micro-US images.
For example, pixels in regions with image artifacts are more challenging to classify. 
To improve image segmentation in those hard-to-segment regions, we propose an AG-BCE loss that penalizes prediction errors in hard regions. The AG-BCE loss is expressed as follows:
\begin{equation}
\begin{split}
     \text{AG-BCE}(P,Y) 
     &= -\frac{1}{K}\sum_{i = 1}^{K} w_i \big( y_i\cdot \log p_i \\
     &+ (1-y_i)\cdot \log (1-p_i) \big )
\end{split}
\end{equation}
where $w_i$ is the weight assigned to pixel $i$.
For pixels in easy regions, $w_i = 1$; for pixels in hard regions, $w_i > 1$.
In this paper, we chose $w_i = 4$ for pixels in hard regions.


\subsubsection{Training loss}

We propose to combine multi-scale deep supervision with the proposed AG-BCE loss.
In our approach, the AG-BCE loss is applied to segmentation predictions generated by all layers.
The loss function for training our MicroSegNet model is defined as follows:
\begin{equation}
\begin{split}
     \text{loss} 
     &= \text{ AG-BCE}(P_4,Y_4) +\text{ AG-BCE}(P_3,Y_3) \\
     &+ \text{ AG-BCE}(P_2,Y_2) + \text{ AG-BCE}(P_1,Y_1)
\end{split}
\end{equation}
 where $P_i$ and $Y_i$ are  the predicted and ground truth segmentations at layer $i$.


\subsection{Experimental design and implementation details}
We used a dataset of 2060 micro-US images from 55 patients for training and 758 images from 20 patients as an independent dataset for evaluation. All images were resized to a dimension of $224\times224$ and normalized to the range $[0, 1]$.
Three segmentation models were utilized for comparison: UNet~\citep{ronneberger2015u}, TransUNet~\citep{chen2021transunet}, and Dilated Residual UNet (DRUNet)~\citep{vesal2022domain}.
During training, we employed an image patch size of $16$, a batch size of $16$, a learning rate of $0.01$ with a momentum of $0.9$, and a weight decay of $1e-4$. 
These hyperparameters were empirically chosen and kept consistent for all models. 
All models were trained for 10 epochs to prevent overfitting.
To ensure robustness, we repeated the training for each model eight times. The average Dice coefficient and Hausdorff distance were computed across all eight runs as the final output for each model.
The training and evaluation processes were performed on a high-performance server equipped with NVIDIA A100 graphical processing units with 80 GB memory. We used Python 3.10.6 and PyTorch 1.12.1 for all experiments.

\subsection{Evaluation metrics}
\label{sect:performanceevaluation}
We used the Dice coefficient to measure the relative overlap between the predicted prostate segmentation $(P)$ and the ground truth prostate segmentation $(G)$, defined as follows:

\begin{equation}
DSC(G, P) = \frac{2 \times |G \cap P|}{|G| + |P|}
\end{equation}

Here, $|G|$ and $|P|$ denote the number of positive pixels in the ground truth and predicted segmentation images, respectively. The intersection of $G$ and $P$ is denoted by $G\cap P$. The Dice coefficient ranges from $0$ to $1$, where a higher score indicates a better overlap between the two segmentations.

In addition to the Dice coefficient, we employed the Hausdorff distance to measure the maximum distance between two prostate boundaries $G$ and $P$, defined as 
\begin{equation}
HD = \max \left\{ \sup_{g \in \mathcal{G}}  \inf_{p \in P} d(g, p), \sup_{p \in \mathcal{P}}  \inf_{g \in \mathcal{G}} d(g, p)  \right\}
\end{equation}
To eliminate the impact of small outliers, we used the 95th percentile (HD95) of the distances between boundary points instead of the maximum distance (HD).

\section{Results}

\subsection{Qualitative results} 
\begin{figure*}[!hbt]
\centering
\includegraphics[width=1\textwidth]{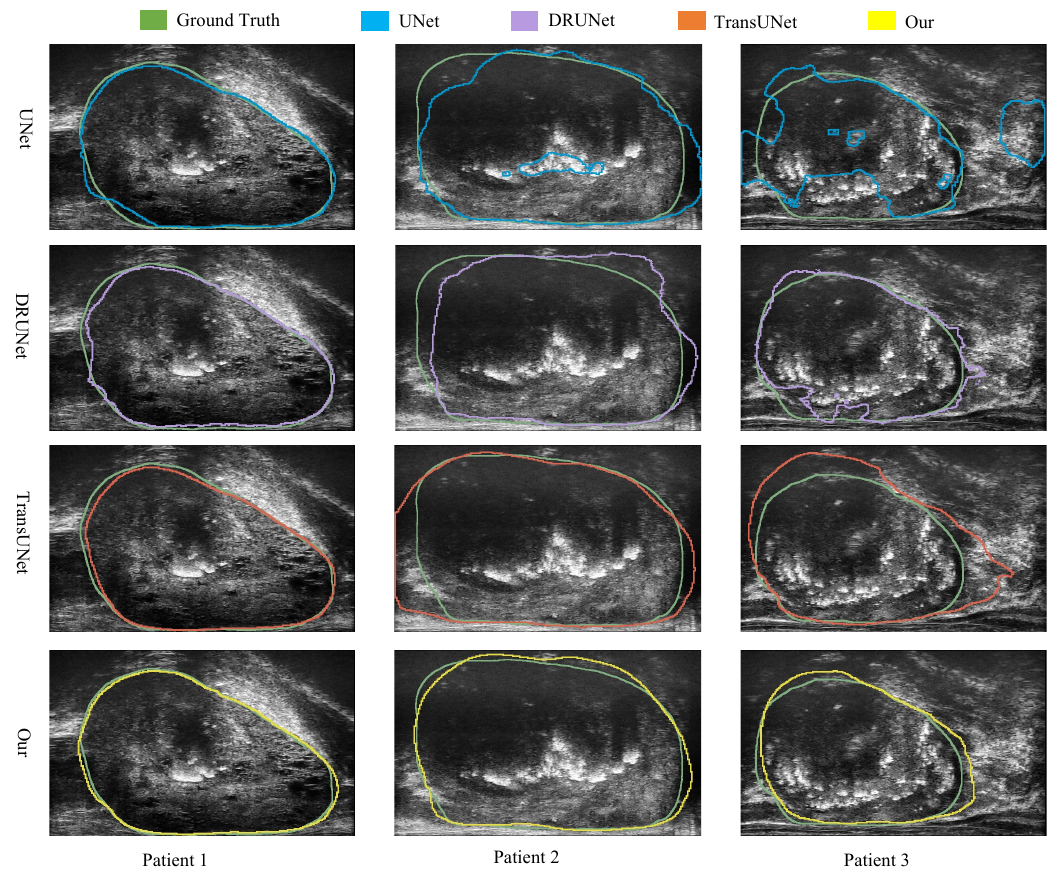}
\caption{Visual prostate segmentation results of three representative patients. Patient 1: relatively clear boundary with rare artifacts. Patient 2: indistinct boundary with some artifacts. Patient 3: indistinct boundary with substantial artifacts.}
\label{fig:segmentation_results}
\end{figure*}

Figure~\ref{fig:segmentation_results} presents the qualitative segmentation results of four different models across three representative patients.
For patient 1, the micro-US scan exhibited high image quality, with clear prostate boundaries and minimal artifacts. As a result, all four models demonstrated accurate prostate segmentation.
In contrast, the micro-US images of patient 2 and patient 3 depicted indistinct prostate boundaries and artifacts caused by prostate calcification. In these challenging-to-segment cases, our MicroSegNet model outperformed the other three models. Notably, the performance of the UNet model significantly deteriorated in the presence of large image artifacts. This can be attributed to the absence of a self-attention mechanism, which hampers its ability to gain a comprehensive understanding of the input images.

Compared to the Dilated UNet and TransUNet models, our MicroSegNet achieved more precise and smoother segmentation of the prostate boundary, showcasing its accuracy and robustness against artifacts. This improvement primarily stems from the integration of multi-scale information and the adoption of the AG-BCE loss, enabling the model to effectively focus on challenging regions during training.

\subsection{Quantitative results}
Table~\ref{method result} presents the Dice coefficients and Hausdorff distances of four segmentation models averaged over 20 testing cases.
Our MicroSegNet outperformed the other three models in terms of both the Dice coefficient and Hausdorff distance.
By incorporating multi-scale deep supervision and employing the AG-BCE loss, we were able to improve the Dice coefficient of TransUNet by 0.007 and reduce the Hausdorff distance by 0.10 mm.
The improvement in the Dice coefficient from 0.932 to 0.939 and in the Hausdorff distance from 2.12 mm to 2.02 mm is small but remains practically significant for two reasons. First, prostate capsule annotation is the most common methodology for registration between ultrasound and other imaging modalities such as MRI and PSMA PET-CT. Therefore, small errors tend to rapidly compound when transferring images between modalities. Second, capsular identification is likely the initial step in identifying whether prostate cancers have extended through the capsule. This is often a very subtle finding on imaging and is poorly represented in MRI and PSMA PET-CT. Thus, highly accurate capsular annotation will pave the way for precision diagnostics and staging in the future.
Notably, the baseline UNet model exhibited significantly poorer performance compared to the TransUNet and MicroSegNet models, which utilize vision transformers. This suggests the advantages of capturing long-range dependencies in challenging image segmentation tasks.
The results presented in Table~\ref{method result} also indicate that the use of the boundary-weighted segmentation loss (BWSL) does not improve the Dice coefficient (0.932 vs. 0.932) or the Hausdorff distance (2.12 mm vs. 2.17 mm).

\begin{table}[!hbt]
  \centering
  \caption{Comparison of different segmentation methods using Dice coefficient and Hausdorff distance (mm).}
  \label{method result}
  \begin{tabular}{cccc}
    \toprule
    &DSC (p-value) &HD95  (p-value)\\
    \midrule
    UNet & 0.910 (<0.01) & 3.26 (<0.01) \\
    DRUNet & 0.914 (<0.01) & 3.09 (<0.01)\\
    TransUnet & 0.932 (<0.01) & 2.12 (0.030) \\
     TransUnet + BWSL  & 0.932 (<0.01) & 2.17 (<0.01)\\
    MicroSegNet (Our) & \bf{0.939} & \bf{2.02} \\
    \bottomrule
  \end{tabular}
\end{table}

Our MicroSegNet model was also evaluated against three human annotators: a master student, a medical student, and a urologist. 
The Dice coefficient and Hausdorff distance for each annotator averaged over 20 testing cases are presented in Table \ref{human result}. 
Our MicroSegNet model exhibited  superior performance compared to  all three human annotators.
This highlights the potential of our segmentation model to assist clinicians in accurately localizing the prostate during micro-US-guided biopsy.

\begin{table}[!hbt]
  \centering
  \caption{Comparison between human annotators and our MicroSegNet model using Dice coefficient (DSC) and Hausdorff distance (mm).}
  \label{human result}
  \begin{tabular}{cccc}
    \toprule
    &DSC (p-value) &HD95 (p-value) \\
    \midrule
    Medical Student & 0.880 (<0.01) & 4.20 (<0.01) \\
    Master Student & 0.923 (<0.01) & 3.03 (<0.01) \\
    Clinician & 0.922 (0.055) & 3.00 (<0.01) \\
    MicroSegNet (Our) & \bf{0.939} & \bf{2.02} \\
    \bottomrule
  \end{tabular}
\end{table}

In addition, we have conducted several statistical tests to confirm the statistical significance of the enhanced Dice coefficient and Hausdorff distance achieved by our MicroSegNet. To compare the performance of our MicroSegNet with other methods or annotators, we consider whether our MicroSegNet achieves higher Dice coefficients and lower Hausdorff distances vs. others. To test these, we conduct the paired t-tests to inform which method achieves a better performance. As presented in Table 1 and Table 2, our MicroSegNet model significantly outperforms (p-values < 0.05) the U-Net, DRUNet, TransUNet segmentation models, and two student annotators.

\subsection{Ablation Study.}
We explored the effect of the ratio of the weights for the hard and easy regions, i.e., {$\frac{W_{hard}}{W_{easy}}$, on the performance of our MicroSegNet model.
Figure \ref{fig:ablation} shows the Dice coefficients and Hausdorff distances of MicroSegNet at different  {$\frac{W_{hard}}{W_{easy}}$ values. 
By gradually increasing the weight ratio between hard and easy regions during the training process, we observed a corresponding improvement in the performance of our model. The optimal weight ratio that yielded peak performance was found to be $\frac{W_{hard}}{W_{easy}} = 4$. At this weight ratio, the model effectively prioritized the hard regions, resulting in enhanced segmentation accuracy.

However, we noticed that as the weight ratio exceeded this optimal value, the model's performance began to decline. This decline was attributed to the model excessively focusing on the hard regions, which caused a deterioration in performance on the easier regions. In essence, the model became overly biased towards the challenging regions, compromising its ability to accurately segment the easier regions.

\begin{figure}[!hbt]
    \centering
    \includegraphics[width=0.48\textwidth]{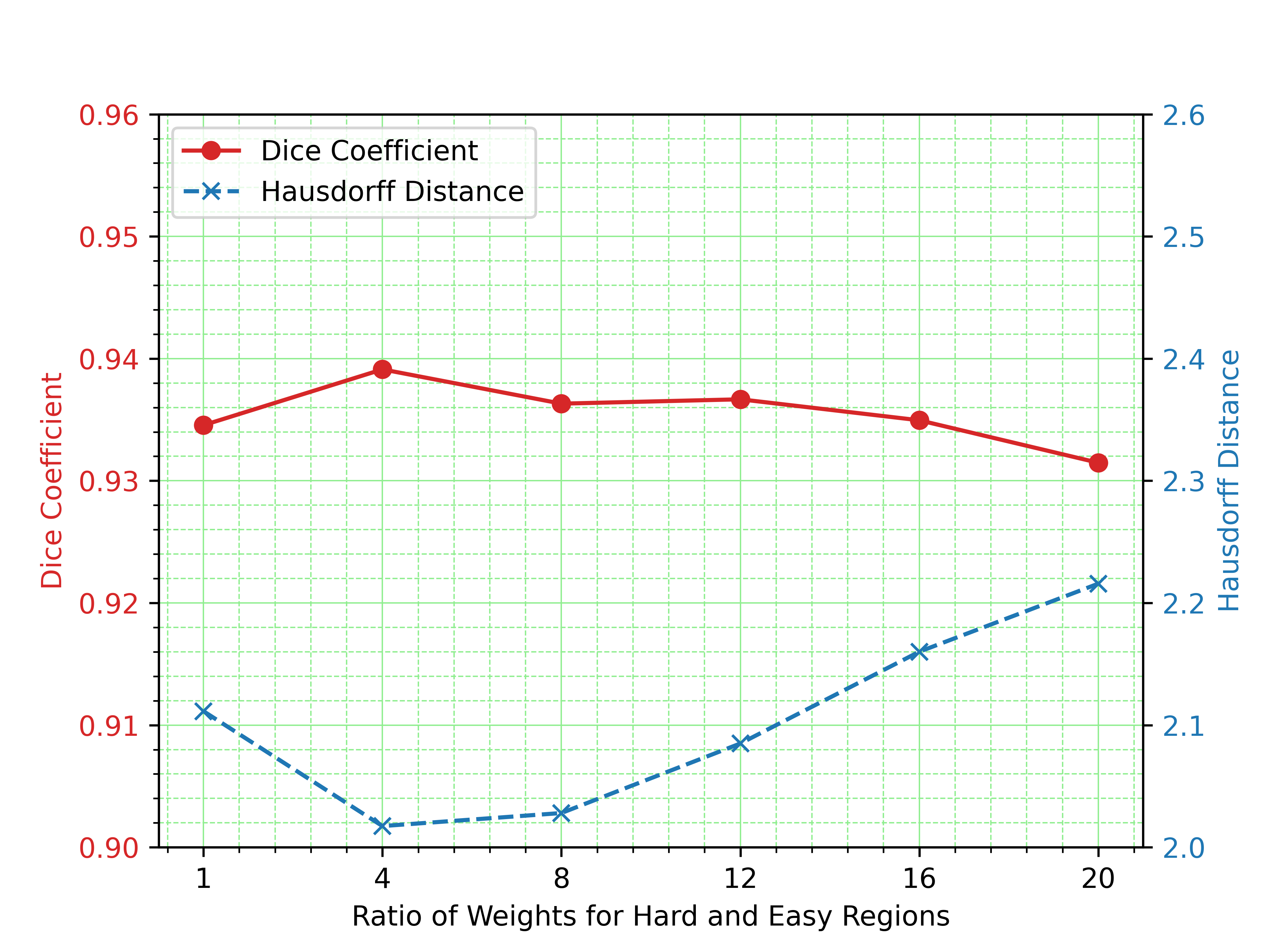}
    \caption{The impact of $\frac{W_{hard}}{W_{easy}}$ on model performance. 
    \label{fig:ablation}}
\end{figure}

\section{Discussion}

The prostate anatomy where the model had the most difficulty was towards the midline of the prostate, where the prostate and bladder fuse to a continuum and distinct prostate boundaries are difficult to determine. 
Segmentation of these areas is difficult even for expert urologists, and it is possible that variations in the model correspond to variations in the ground truth annotations from the annotators. 
We expect better performance of our model as we include more training cases so that midline variations between expert segmentations can be mitigated.

\subsection{Clinical implications}
Micro-US has emerged as an inexpensive technology for real-time prostate cancer imaging.  Additionally, the accuracy of prostate cancer detection and local clinical staging by micro-US appears equivalent or superior compared with MRI \citep{lughezzani2021diagnostic}.  However, a major limitation of micro-US, common to all ultrasound modalities, is the lack of specific 3D orientation for imaging data.  Because micro-US images are acquired in a 'half-doughnut' shaped image stack, the voxels do not conform to the conventional DICOM format.  Despite the promising accuracy of micro-US, the lack of 3D orientation prevents accurate tracking of biopsy sample location, comparison to other imaging modalities, and complex treatment planning.

Additionally, a well-designed comparative study between conventional and micro-US suggests that interpretation of micro-US is challenging \citep{pavlovich2021multi}.  The capsular annotation proposed in our manuscript is an important initial step towards automated cancer detection by establishing the ability to compare micro-US to biopsy samples and surgical pathology at scale.  Finally, the automated capsular annotation could provide a point-of-care image quality assessment, thus enabling the standardization of image acquisition across multiple providers.  

\subsection{Limitations and strengths}
Our results must be interpreted within the limitations of the imaging data and the study design.  The images were collected at a single institution by a single surgeon, which may introduce bias.  Furthermore, our study included patients with extracapsular extension (ECE). However, it is important to emphasize that these cases involved only microscopic ECE. Notably, none of our patients had T4 disease, which is characterized by the cancer grossly invading other structures.

The micro-US device and image settings were obtained using the manufacturer presets in order to provide uniformity.  Micro-US is an excellent imaging modality for machine learning implementation within the clinical setting as device manufacturing is performed by a single company with standard probes and imaging presets.  Additionally, each device contains a graphical processing unit capable of completing heavy computational tasks.  The images for our data set were obtained under routine clinical practice and were exactingly reviewed by a clinical expert.  Thus, our data represents one of the best representative yet cleanest imaging datasets currently available for micro-US.

\section{Conclusion}
This paper presents the first deep learning approach for automated segmentation of the prostate capsule on micro-ultrasound images. 
One of our key contributions is the introduction of an AG-BCE loss, which incorporates non-expert annotations for improved learning in challenging segmentation regions typically difficult for non-experts.
The results obtained demonstrate the superior performance of our approach compared to other state-of-the-art methods, highlighting the effectiveness of the proposed loss. Our segmentation tool also has the potential to assist urologists in prostate biopsy and cancer diagnosis, offering valuable support in clinical decision-making.
To advance research in this field, we will open-source the micro-ultrasound dataset, inclusive of both expert and non-expert prostate annotations, to stimulate further investigations in automated prostate segmentation on micro-ultrasound images.

\printcredits

\appendix

\section*{APPENDIX}
Table \ref{tab:quantativeresults_dice} and Table \ref{table_hd} present the Dice coefficient and Hausdorff distance, respectively, for each of the 20 test cases across the four segmentation models.
\begin{table}[!h]                           
	\centering
 \caption{Dice Coefficients of 20 testing cases.}
	\renewcommand{\arraystretch}{1.3} 
 \scalebox{0.83}{
	\begin{tabular}{>{\centering\arraybackslash}p{0.7cm}|>{\centering\arraybackslash}p{1.3cm}|>{\centering\arraybackslash}p{1.3cm}|>{\centering\arraybackslash}p{1.3cm}|>{\centering\arraybackslash}p{1.3cm}|>{\centering\arraybackslash}p{1.3cm}}
		\toprule[1.5pt]
  Case & \multirow{2}{*}{UNet} & Dilated & Trans & TransUNet & MicroSegNet \\
  No. & & UNet & UNet & BWSL & (Ours)\\
		\midrule[1.5pt]
        1	&	0.921	&	0.926	&	0.932	&	0.933	&	0.934   \\
        2	&	0.923	&	0.921	&	0.944	&	0.948	&	0.949   \\
        3	&	0.943	&	0.947	&	0.954	&	0.954	&	0.955   \\
        4	&	0.917	&	0.929	&	0.943	&	0.936	&	0.944   \\
        5	&	0.940	&	0.936	&	0.946	&	0.949	&	0.953   \\
        6	&	0.833	&	0.851	&	0.893	&	0.887	&	0.909   \\
        7	&	0.957	&	0.943	&	0.955	&	0.954	&	0.958   \\
        8	&	0.929	&	0.926	&	0.940	&	0.944	&	0.952   \\
        9	&	0.821	&	0.829	&	0.918	&	0.908	&	0.923   \\
        10	&	0.843	&	0.865	&	0.904	&	0.907	&	0.904   \\
        11	&	0.889	&	0.898	&	0.927	&	0.922	&	0.930   \\
        12	&	0.865	&	0.919	&	0.904	&	0.911	&	0.940   \\
        13	&	0.901	&	0.885	&	0.937	&	0.936	&	0.945   \\
        14	&	0.950	&	0.953	&	0.958	&	0.956	&	0.958   \\
        15	&	0.956	&	0.952	&	0.950	&	0.945	&	0.946   \\
        16	&	0.922	&	0.931	&	0.948	&	0.945	&	0.947   \\
        17	&	0.938	&	0.941	&	0.945	&	0.943	&	0.949   \\
        18	&	0.948	&	0.947	&	0.954	&	0.957	&	0.959   \\
        19	&	0.929	&	0.926	&	0.927	&	0.928	&	0.928   \\
        20	&	0.872	&	0.850	&	0.866	&	0.877	&	0.897   \\
        \hline
        Mean & 0.910 & 0.914 & 0.932	&  0.932     & \textbf{0.939} \\
		\bottomrule[1.5pt]
	\end{tabular}}
	\label{tab:quantativeresults_dice}                           	
\end{table}

\begin{table}[!h]                           
	\centering
 \caption{Hausdorff distances (mm) of 20 testing cases.}
	\renewcommand{\arraystretch}{1.3} 
	\scalebox{0.83}{
	\begin{tabular}{>{\centering\arraybackslash}p{0.7cm}|>{\centering\arraybackslash}p{1.3cm}|>{\centering\arraybackslash}p{1.3cm}|>{\centering\arraybackslash}p{1.3cm}|>{\centering\arraybackslash}p{1.3cm}|>{\centering\arraybackslash}p{1.3cm}}
		\toprule[1.5pt]
  Case & \multirow{2}{*}{UNet} & Dilated & Trans & TransUNet & MicroSegNet \\
  No. & & UNet & UNet & BWSL & (Ours)\\
		\midrule[1.5pt]
        1	&	2.17	&	2.12	&	1.82	&	1.84	&	 1.79  \\
        2	&	1.96	&	2.07	&	1.16	&	1.11	&	 1.08  \\
        3	&	2.06	&	1.81	&	1.54	&	1.52	&	 1.48  \\
        4	&	3.57	&	2.64	&	1.90	&	2.13	&	 1.85  \\
        5	&	2.82	&	2.86	&	1.93	&	1.81	&	 1.74  \\
        6	&	5.40	&	3.70	&	2.67	&	2.97	&	 2.53  \\
        7	&	1.58	&	2.58	&	1.44	&	1.50	&	 1.31  \\
        8	&	2.10	&	2.33	&	1.72	&	1.54	&	 1.43  \\
        9	&	5.65	&	5.34	&	2.96	&	2.97	&	 2.69  \\
        10	&	5.06	&	4.81	&	2.86	&	2.86	&	 2.87 \\
        11	&	5.64	&	5.45	&	3.05	&	3.01	&	 3.01  \\
        12	&	3.23	&	3.21	&	2.73	&	2.64	&	 2.37  \\
        13	&	3.43	&	3.61	&	2.34	&	2.35	&	 1.96  \\
        14	&	1.97	&	1.84	&	1.39	&	1.41	&	 1.38  \\
        15	&	1.63	&	1.75	&	2.26	&	2.78	&	 2.77  \\
        16	&	4.60	&	3.89	&	2.49	&	2.81	&	 2.62  \\
        17	&	3.06	&	2.55	&	2.11	&	2.13	&	 1.92  \\
        18	&	1.71	&	1.94	&	1.22	&	1.23	&	 1.10  \\
        19	&	2.50	&	2.39	&	2.04	&	2.01	&	 1.86  \\
        20	&	5.02	&	4.94	&	2.84	&	2.81	&	 2.65  \\
        \hline
        Mean & 3.26 & 3.09 & 2.12	&   2.17   &  2.02\\
		\bottomrule[1.5pt]
	\end{tabular}}
	\label{table_hd}                           	
\end{table}

We investigate the differences between expert and student annotators with respect to the rotation angles of micro-ultrasound images, where 0 degrees roughly corresponds to the midline. For each 2D micro-US slice, we computed the Hausdorff distance (in pixels) between its expert and non-expert annotations. Figure \ref{fig:diff_angle} shows the histogram of the Hausdorff distances across different rotation angles (averaged over 55 training cases). We can observe that the difference between the two sets of annotations is largest near the midline and decreases as we move toward the lateral sides of the prostate. 

\begin{figure}[!hbt]
    \centering
    \includegraphics[width=0.48\textwidth]{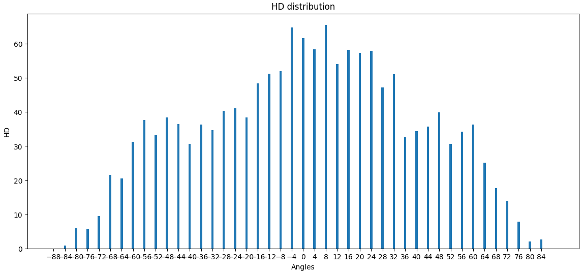}
    \caption{The difference between expert and non-expert annotations with respect to rotation angle. 
    \label{fig:diff_angle}}
\end{figure}

Table \ref{table_clinicalsignificance} provides additional information on the patients, including Prostate-Specific Antigen (PSA), age, prostate volume, number of patients with clinically significant cancer, and extracapsular extension for the training, testing, and entire groups.

\begin{table}[!hbt]                           
	\centering
 \caption{Additional patient information.}
	\renewcommand{\arraystretch}{1.3} 
	\scalebox{0.85}{
	\begin{tabular}{>{\centering\arraybackslash}p{0.7cm}|>{\centering\arraybackslash}p{1.3cm}|>{\centering\arraybackslash}p{1.3cm}|>{\centering\arraybackslash}p{1.3cm}|>{\centering\arraybackslash}p{1.3cm}|>{\centering\arraybackslash}p{1.3cm}}
		\toprule[1.5pt]
        \multirow{3}{*}{\rotatebox[origin=c]{90}{Dataset}} & \multirow{2}{*}{PSA } & \multirow{2}{*}{Age}   & Prostate & Clinically  & Extra- \\
  &  (ng/mL) & (Years) &  Volume &Significant  & Capsular \\
  &   &  &   (ml) &Cancer & Extension\\
		\midrule[1.5pt]
        Train	&	9.6	&	68.9	&	52.2	&	63.6\%	&	 20\%  \\
        Test	&	7.1	&	67.5	&	42.3	&	50.0\%	&	 0 \% \\
        Entire	&	8.9	&	68.5	&	49.7	&	60.0\%	&	  14.7\% \\
		\bottomrule[1.5pt]
	\end{tabular}}
	\label{table_clinicalsignificance}                           	
\end{table}

\section*{Conflict of interest statement}
The authors have no conflict of interest to declare.

\section*{Declaration of generative AI in scientific writing}
During the preparation of this work the authors used the ChatGPT-3.5 model in order to improve the readability and language of our paper. After using this tool/service, the authors reviewed and edited the content as needed and take full responsibility for the content of the publication.

\section*{Acknowledgments}
This work was supported by the Department of Medicine and the Intelligent Critical Care Center at the University of Florida College of Medicine.
We would like to express our gratitude to Jessica Kirwan for editing the language of this paper.

\bibliographystyle{apalike}


\bibliography{biblioentropy} 

\begin{thebibliography}{}

\bibitem[Brisbane et~al., 2022]{brisbane2022targeted}
Brisbane, W.~G., Priester, A.~M., Ballon, J., Kwan, L., Delfin, M.~K., Felker, E.~R., Sisk, A.~E., Hu, J.~C., and Marks, L.~S. (2022).
\newblock Targeted prostate biopsy: umbra, penumbra, and value of perilesional sampling.
\newblock {\em European urology}.

\bibitem[Chen et~al., 2021a]{chen2021transunet}
Chen, J., Lu, Y., Yu, Q., Luo, X., Adeli, E., Wang, Y., Lu, L., Yuille, A.~L., and Zhou, Y. (2021a).
\newblock Transunet: Transformers make strong encoders for medical image segmentation.
\newblock {\em arXiv preprint arXiv:2102.04306}.

\bibitem[Chen et~al., 2021b]{chen2021medical}
Chen, J., Wan, Z., Zhang, J., Li, W., Chen, Y., Li, Y., and Duan, Y. (2021b).
\newblock Medical image segmentation and reconstruction of prostate tumor based on 3d alexnet.
\newblock {\em Computer methods and programs in biomedicine}, 200:105878.

\bibitem[Dias et~al., 2022]{dias2022multiparametric}
Dias, A.~B., O’Brien, C., Correas, J.-M., and Ghai, S. (2022).
\newblock Multiparametric ultrasound and micro-ultrasound in prostate cancer: a comprehensive review.
\newblock {\em The British Journal of Radiology}, 95(1131):20210633.

\bibitem[Ding et~al., 2023]{ding2023multi}
Ding, M., Lin, Z., Lee, C.~H., Tan, C.~H., and Huang, W. (2023).
\newblock A multi-scale channel attention network for prostate segmentation.
\newblock {\em IEEE Transactions on Circuits and Systems II: Express Briefs}.

\bibitem[Dosovitskiy et~al., 2020]{dosovitskiy2020image}
Dosovitskiy, A., Beyer, L., Kolesnikov, A., Weissenborn, D., Zhai, X., Unterthiner, T., Dehghani, M., Minderer, M., Heigold, G., Gelly, S., et~al. (2020).
\newblock An image is worth 16x16 words: Transformers for image recognition at scale.
\newblock {\em arXiv preprint arXiv:2010.11929}.

\bibitem[Fedorov et~al., 2012]{fedorov20123d}
Fedorov, A., Beichel, R., Kalpathy-Cramer, J., Finet, J., Fillion-Robin, J.-C., Pujol, S., Bauer, C., Jennings, D., Fennessy, F., Sonka, M., et~al. (2012).
\newblock 3d slicer as an image computing platform for the quantitative imaging network.
\newblock {\em Magnetic resonance imaging}, 30(9):1323--1341.

\bibitem[Ghavami et~al., 2019]{ghavami2019automatic}
Ghavami, N., Hu, Y., Gibson, E., Bonmati, E., Emberton, M., Moore, C.~M., and Barratt, D.~C. (2019).
\newblock Automatic segmentation of prostate mri using convolutional neural networks: Investigating the impact of network architecture on the accuracy of volume measurement and mri-ultrasound registration.
\newblock {\em Medical image analysis}, 58:101558.

\bibitem[Hatamizadeh et~al., 2022]{hatamizadeh2022unetr}
Hatamizadeh, A., Tang, Y., Nath, V., Yang, D., Myronenko, A., Landman, B., Roth, H.~R., and Xu, D. (2022).
\newblock Unetr: Transformers for 3d medical image segmentation.
\newblock In {\em Proceedings of the IEEE/CVF winter conference on applications of computer vision}, pages 574--584.

\bibitem[Hung et~al., 2022]{hung2022cat}
Hung, A. L.~Y., Zheng, H., Miao, Q., Raman, S.~S., Terzopoulos, D., and Sung, K. (2022).
\newblock Cat-net: A cross-slice attention transformer model for prostate zonal segmentation in mri.
\newblock {\em IEEE Transactions on Medical Imaging}, 42(1):291--303.

\bibitem[Jia et~al., 2017]{jia2017prostate}
Jia, H., Xia, Y., Cai, W., Fulham, M., and Feng, D.~D. (2017).
\newblock Prostate segmentation in mr images using ensemble deep convolutional neural networks.
\newblock In {\em 2017 IEEE 14th international symposium on biomedical imaging (ISBI 2017)}, pages 762--765. IEEE.

\bibitem[Karimi and Salcudean, 2019]{karimi2019reducing}
Karimi, D. and Salcudean, S.~E. (2019).
\newblock Reducing the hausdorff distance in medical image segmentation with convolutional neural networks.
\newblock {\em IEEE Transactions on medical imaging}, 39(2):499--513.

\bibitem[Karimi et~al., 2019]{KARIMI2019186}
Karimi, D., Zeng, Q., Mathur, P., Avinash, A., Mahdavi, S., Spadinger, I., Abolmaesumi, P., and Salcudean, S.~E. (2019).
\newblock Accurate and robust deep learning-based segmentation of the prostate clinical target volume in ultrasound images.
\newblock {\em Medical Image Analysis}, 57:186--196.

\bibitem[Kervadec et~al., 2019]{kervadec2019boundary}
Kervadec, H., Bouchtiba, J., Desrosiers, C., Granger, E., Dolz, J., and Ayed, I.~B. (2019).
\newblock Boundary loss for highly unbalanced segmentation.
\newblock In {\em International conference on medical imaging with deep learning}, pages 285--296. PMLR.

\bibitem[Klotz et~al., 2021]{klotz2021comparison}
Klotz, L., Lughezzani, G., Maffei, D., S{\'a}nchez, A., Pereira, J.~G., Staerman, F., Cash, H., Luger, F., Lopez, L., Sanchez-Salas, R., et~al. (2021).
\newblock Comparison of micro-ultrasound and multiparametric magnetic resonance imaging for prostate cancer: A multicenter, prospective analysis.
\newblock {\em Canadian Urological Association Journal}, 15(1):E11.

\bibitem[Lei et~al., 2019a]{lei2019ultrasound}
Lei, Y., Tian, S., He, X., Wang, T., Wang, B., Patel, P., Jani, A.~B., Mao, H., Curran, W.~J., Liu, T., et~al. (2019a).
\newblock Ultrasound prostate segmentation based on multidirectional deeply supervised v-net.
\newblock {\em Medical physics}, 46(7):3194--3206.

\bibitem[Lei et~al., 2019b]{lei2019ultrasound_2}
Lei, Y., Wang, T., Wang, B., He, X., Tian, S., Jani, A.~B., Mao, H., Curran, W.~J., Patel, P., Liu, T., et~al. (2019b).
\newblock Ultrasound prostate segmentation based on 3d v-net with deep supervision.
\newblock In {\em Medical Imaging 2019: Ultrasonic Imaging and Tomography}, volume 10955, pages 198--204. SPIE.

\bibitem[Loeb et~al., 2011]{loeb2011complications}
Loeb, S., Carter, H.~B., Berndt, S.~I., Ricker, W., and Schaeffer, E.~M. (2011).
\newblock Complications after prostate biopsy: data from seer-medicare.
\newblock {\em The Journal of urology}, 186(5):1830--1834.

\bibitem[Lughezzani et~al., 2021]{lughezzani2021diagnostic}
Lughezzani, G., Maffei, D., Saita, A., Paciotti, M., Diana, P., Buffi, N.~M., Colombo, P., Elefante, G.~M., Hurle, R., Lazzeri, M., et~al. (2021).
\newblock Diagnostic accuracy of microultrasound in patients with a suspicion of prostate cancer at magnetic resonance imaging: a single-institutional prospective study.
\newblock {\em European urology focus}, 7(5):1019--1026.

\bibitem[Ma et~al., 2020]{ma2020distance}
Ma, J., Wei, Z., Zhang, Y., Wang, Y., Lv, R., Zhu, C., Gaoxiang, C., Liu, J., Peng, C., Wang, L., et~al. (2020).
\newblock How distance transform maps boost segmentation cnns: an empirical study.
\newblock In {\em Medical Imaging with Deep Learning}, pages 479--492. PMLR.

\bibitem[Mottet et~al., 2021]{mottet2021eau}
Mottet, N., van~den Bergh, R.~C., Briers, E., Van~den Broeck, T., Cumberbatch, M.~G., De~Santis, M., Fanti, S., Fossati, N., Gandaglia, G., Gillessen, S., et~al. (2021).
\newblock Eau-eanm-estro-esur-siog guidelines on prostate cancer—2020 update. part 1: screening, diagnosis, and local treatment with curative intent.
\newblock {\em European urology}, 79(2):243--262.

\bibitem[Orlando et~al., 2020]{https://doi.org/10.1002/mp.14134}
Orlando, N., Gillies, D.~J., Gyacskov, I., Romagnoli, C., D’Souza, D., and Fenster, A. (2020).
\newblock Automatic prostate segmentation using deep learning on clinically diverse 3d transrectal ultrasound images.
\newblock {\em Medical Physics}, 47(6):2413--2426.

\bibitem[Pavlovich et~al., 2021]{pavlovich2021multi}
Pavlovich, C., Hyndman, M., Eure, G., Ghai, S., Caumartin, Y., Herget, E., Young, J., Wiseman, D., Caughlin, C., Gray, R., et~al. (2021).
\newblock A multi-institutional randomized controlled trial comparing first-generation transrectal high-resolution micro-ultrasound with conventional frequency transrectal ultrasound for prostate biopsy.
\newblock {\em BJUI compass}, 2(2):126--133.

\bibitem[Peng et~al., 2022a]{peng2022h}
Peng, T., Tang, C., Wu, Y., and Cai, J. (2022a).
\newblock H-segmed: a hybrid method for prostate segmentation in trus images via improved closed principal curve and improved enhanced machine learning.
\newblock {\em International Journal of Computer Vision}, 130(8):1896--1919.

\bibitem[Peng et~al., 2022b]{PENG2022106752}
Peng, T., Wu, Y., Qin, J., Wu, Q.~J., and Cai, J. (2022b).
\newblock H-proseg: Hybrid ultrasound prostate segmentation based on explainability-guided mathematical model.
\newblock {\em Computer Methods and Programs in Biomedicine}, 219:106752.

\bibitem[Peng et~al., 2023]{peng2023automatic}
Peng, T., Xu, D., Tang, C., Zhao, J., Shen, Y., Yang, C., and Cai, J. (2023).
\newblock Automatic coarse-to-refinement-based ultrasound prostate segmentation using optimal polyline segment tracking method and deep learning.
\newblock {\em Applied Intelligence}, pages 1--17.

\bibitem[Ronneberger et~al., 2015]{ronneberger2015u}
Ronneberger, O., Fischer, P., and Brox, T. (2015).
\newblock U-net: Convolutional networks for biomedical image segmentation.
\newblock In {\em International Conference on Medical image computing and computer-assisted intervention}, pages 234--241. Springer.

\bibitem[Siegel et~al., 2022]{siegel2022cancer}
Siegel, R.~L., Miller, K.~D., Fuchs, H.~E., and Jemal, A. (2022).
\newblock Cancer statistics, 2022.
\newblock {\em CA: a cancer journal for clinicians}.

\bibitem[Soerensen et~al., 2021]{soerensen2021prognet}
Soerensen, S. J.~C., Fan, R., Seetharaman, A., Chen, L., Shao, W., Bhattacharya, I., Borre, M., Chung, B., To’o, K., Sonn, G., et~al. (2021).
\newblock Prognet: prostate gland segmentation on mri with deep learning.
\newblock In {\em Medical Imaging 2021: Image Processing}, volume 11596, pages 743--750. SPIE.

\bibitem[Sountoulides et~al., 2021]{sountoulides2021micro}
Sountoulides, P., Pyrgidis, N., Polyzos, S.~A., Mykoniatis, I., Asouhidou, E., Papatsoris, A., Dellis, A., Anastasiadis, A., Lusuardi, L., and Hatzichristou, D. (2021).
\newblock Micro-ultrasound--guided vs multiparametric magnetic resonance imaging-targeted biopsy in the detection of prostate cancer: a systematic review and meta-analysis.
\newblock {\em The Journal of urology}, 205(5):1254--1262.

\bibitem[Vaswani et~al., 2017]{vaswani2017attention}
Vaswani, A., Shazeer, N., Parmar, N., Uszkoreit, J., Jones, L., Gomez, A.~N., Kaiser, {\L}., and Polosukhin, I. (2017).
\newblock Attention is all you need.
\newblock {\em Advances in neural information processing systems}, 30.

\bibitem[Vesal et~al., 2022]{vesal2022domain}
Vesal, S., Gayo, I., Bhattacharya, I., Natarajan, S., Marks, L.~S., Barratt, D.~C., Fan, R.~E., Hu, Y., Sonn, G.~A., and Rusu, M. (2022).
\newblock Domain generalization for prostate segmentation in transrectal ultrasound images: A multi-center study.
\newblock {\em Medical Image Analysis}, 82:102620.

\bibitem[Wang et~al., 2022]{wang2022multiscale}
Wang, B., Wang, {\textperiodcentered}.~F., Dong, P., and Li, {\textperiodcentered}.~C. (2022).
\newblock Multiscale transunet++: dense hybrid u-net with transformer for medical image segmentation.
\newblock {\em Signal, Image and Video Processing}, 16(6):1607--1614.

\bibitem[Wang et~al., 2023]{wang2023prostate}
Wang, X., Chang, Z., Zhang, Q., Li, C., Miao, F., and Gao, G. (2023).
\newblock Prostate ultrasound image segmentation based on dsu-net.
\newblock {\em Biomedicines}, 11(3):646.

\bibitem[Wang et~al., 2019]{8698868}
Wang, Y., Dou, H., Hu, X., Zhu, L., Yang, X., Xu, M., Qin, J., Heng, P.-A., Wang, T., and Ni, D. (2019).
\newblock Deep attentive features for prostate segmentation in 3d transrectal ultrasound.
\newblock {\em IEEE Transactions on Medical Imaging}, 38(12):2768--2778.

\bibitem[Williams et~al., 2022]{williams2022does}
Williams, C., Ahdoot, M., Daneshvar, M.~A., Hague, C., Wilbur, A.~R., Gomella, P.~T., Shih, J., Khondakar, N., Yerram, N., Mehralivand, S., et~al. (2022).
\newblock Why does magnetic resonance imaging-targeted biopsy miss clinically significant cancer?
\newblock {\em The Journal of Urology}, 207(1):95--107.

\bibitem[Wolf et~al., 2010]{wolf2010american}
Wolf, A.~M., Wender, R.~C., Etzioni, R.~B., Thompson, I.~M., D'Amico, A.~V., Volk, R.~J., Brooks, D.~D., Dash, C., Guessous, I., Andrews, K., et~al. (2010).
\newblock American cancer society guideline for the early detection of prostate cancer: update 2010.
\newblock {\em CA: a cancer journal for clinicians}, 60(2):70--98.

\bibitem[Xie and Tu, 2015]{xie2015holistically}
Xie, S. and Tu, Z. (2015).
\newblock Holistically-nested edge detection.
\newblock In {\em Proceedings of the IEEE international conference on computer vision}, pages 1395--1403.

\bibitem[Xie et~al., 2021]{xie2021cotr}
Xie, Y., Zhang, J., Shen, C., and Xia, Y. (2021).
\newblock Cotr: Efficiently bridging cnn and transformer for 3d medical image segmentation.
\newblock In {\em Medical Image Computing and Computer Assisted Intervention--MICCAI 2021: 24th International Conference, Strasbourg, France, September 27--October 1, 2021, Proceedings, Part III 24}, pages 171--180. Springer.

\bibitem[Xu et~al., 2021]{xu2021shadow}
Xu, X., Sanford, T., Turkbey, B., Xu, S., Wood, B.~J., and Yan, P. (2021).
\newblock Shadow-consistent semi-supervised learning for prostate ultrasound segmentation.
\newblock {\em IEEE Transactions on Medical Imaging}, 41(6):1331--1345.

\bibitem[Yang et~al., 2017]{yang2017fine}
Yang, X., Yu, L., Wu, L., Wang, Y., Ni, D., Qin, J., and Heng, P.-A. (2017).
\newblock Fine-grained recurrent neural networks for automatic prostate segmentation in ultrasound images.
\newblock In {\em Proceedings of the AAAI Conference on Artificial Intelligence}, volume~31.

\bibitem[Zhang et~al., 2021]{zhang2021transfuse}
Zhang, Y., Liu, H., and Hu, Q. (2021).
\newblock Transfuse: Fusing transformers and cnns for medical image segmentation.
\newblock In {\em Medical Image Computing and Computer Assisted Intervention--MICCAI 2021: 24th International Conference, Strasbourg, France, September 27--October 1, 2021, Proceedings, Part I 24}, pages 14--24. Springer.

\bibitem[Zhu et~al., 2019]{zhu2019boundary}
Zhu, Q., Du, B., and Yan, P. (2019).
\newblock Boundary-weighted domain adaptive neural network for prostate mr image segmentation.
\newblock {\em IEEE transactions on medical imaging}, 39(3):753--763.

\end{thebibliography}

\end{document}